\documentclass[10pt,conference,a4paper]{IEEEtran}

\usepackage[utf8]{inputenc}
\usepackage{url}

\usepackage{times}

\usepackage{graphicx}
\usepackage{subfigure}
\usepackage{float}
\DeclareGraphicsExtensions{.png,.eps,.ps,.pdf}

\usepackage{url}

\begin{document}

\title{Classification of Spam Emails through Hierarchical Clustering and Supervised Learning}

\author{\IEEEauthorblockN{\small Francisco Jáñez-Martino}
\IEEEauthorblockA{\small Dept. IESA.\\
Universidad de León\\
Researcher at INCIBE\\
fjanm@unileon.es}
\and
\IEEEauthorblockN{\small Eduardo Fidalgo}
\IEEEauthorblockA{\small Dept. IESA.\\
Universidad de León\\
Researcher at INCIBE\\
eduardo.fidalgo@unileon.es}
\and
\IEEEauthorblockN{\small Santiago González-Martínez}
\IEEEauthorblockA{\small Spanish National \\ Cybersecurity \\ Institute (INCIBE)\\
santiago.gonzalez@incibe.es}
\and
\IEEEauthorblockN{\small Javier Velasco-Mata}
\IEEEauthorblockA{\small Dept. IESA.\\
Universidad de León\\
Researcher at INCIBE\\
jvelm@unileon.es}}

\maketitle

\begin{abstract}
Spammers take advantage of email popularity to send indiscriminately unsolicited emails. Although researchers and organizations continuously develop anti-spam filters based on binary classification, spammers bypass them through new strategies, like word obfuscation or image-based spam. For the first time in literature, we propose to classify spam email in categories to improve the handle of already detected spam emails, instead of just using a binary model. First, we applied a hierarchical clustering algorithm to create SPEMC-$11$K (SPam EMail Classification), the first multi-class dataset, which contains three types of spam emails: Health and Technology, Personal Scams, and Sexual Content. Then, we used SPEMC-$11$K to evaluate the combination of TF-IDF and BOW encodings with Naïve Bayes, Logistic Regression and SVM classifiers. Finally, we recommend for the task of multi-class spam classification the use of (i) TF-IDF combined with SVM for the best micro F1 score performance, $95.39\%$, and (ii) TD-IDF along with NB for the fastest spam classification, analyzing an email in $2.13$ms.  
\end{abstract}

\begin{IEEEkeywords}
Spam Email Detection, Multi-classification, Unsupervised Learning, Hierarchical Clustering, Text Classification     
\end{IEEEkeywords}

{\bf Type of contribution:}  {\it Research in development}

\section{Introduction}

Email services are one of the most popular communication media due to its efficiency and quickness. They allow to send and receive messages via the Internet, usually through free and anonymous registration. However, these features eases the bulk and unsolicited emails, best-know as spam. Users indiscriminately receive spam, whose content can include advertisements, digital marketing, or frauds such as malware distribution, leaked-data, and phishing \cite{Ferrara_2019_HistoryDigitalSpam}. Spam emails produce a loss of work productivity and traffic congestion \cite{Fronzetti_2019_MeasuringImpactSpammers}, and spam with a fraudulent aim also risk the security and privacy of who receive them. 

The relatively low cost, straightforward creation and the user's difficulty of identification makes spam emails one of the most used attack vectors for cybercriminals. Considering the reports of Cisco Talos\footnote{\protect\url{https://talosintelligence.com/reputation_center/email_rep} Retrieved March 2020} and Kaspersky Lab\footnote{\protect\url{https://www.statista.com/statistics/420391/spam-email-traffic-share/} Retrieved March 2020}, spam emails represent approximately between $55\%$ and $85\%$ of the daily total volume of worldwide emails. 

The main tools to detect spam are binary filters, i.e. algorithms that categorise emails in spam or not spam. Traditionally, anti-spam filters use manual analysis, pattern matching, and Artificial Intelligence (AI) techniques (\cite{Ferrara_2019_HistoryDigitalSpam, Bhowmick_2018_SpamReviewLiterature}). More recent approaches, mainly based on Automatic Text Classification, provides more accurate models \cite{Bhowmick_2018_SpamReviewLiterature}. Nevertheless, despite the efforts of organisations and researchers to develop more efficient binary spam filters, spammers bypass them \cite{Ferrara_2019_HistoryDigitalSpam}. 

We propose for the first time to the best of our knowledge, to enhance the detection of spam by using an automatic multi-classification approach, instead of a binary model. The multi-classification of spam emails can improve the efficiency in cybersecurity incident handling, companies and citizens protection and early warning by detecting spam campaigns related to specific targets and patterns inside categories. Dada et al. \cite{Dada_2019_MLReview} identified in their literature review that some researchers used the behavioural patterns of spammers as a critical aspect of spam detection. Besides, due to the high daily volume and variety of spam, its detection can be addressed as a Big Data problem that increases the need of solutions that offer an adequate protection service for citizens, companies and response teams. Redmiles et al. \cite{Redmiles_2018_WhoClicksSpam} proposed to help users by identifying what content of the email is suspicious to prevent the troubles occurred by spam messages.

\begin{figure*}[ht]
    \centering
    \includegraphics[width=\textwidth, height=4.5cm]{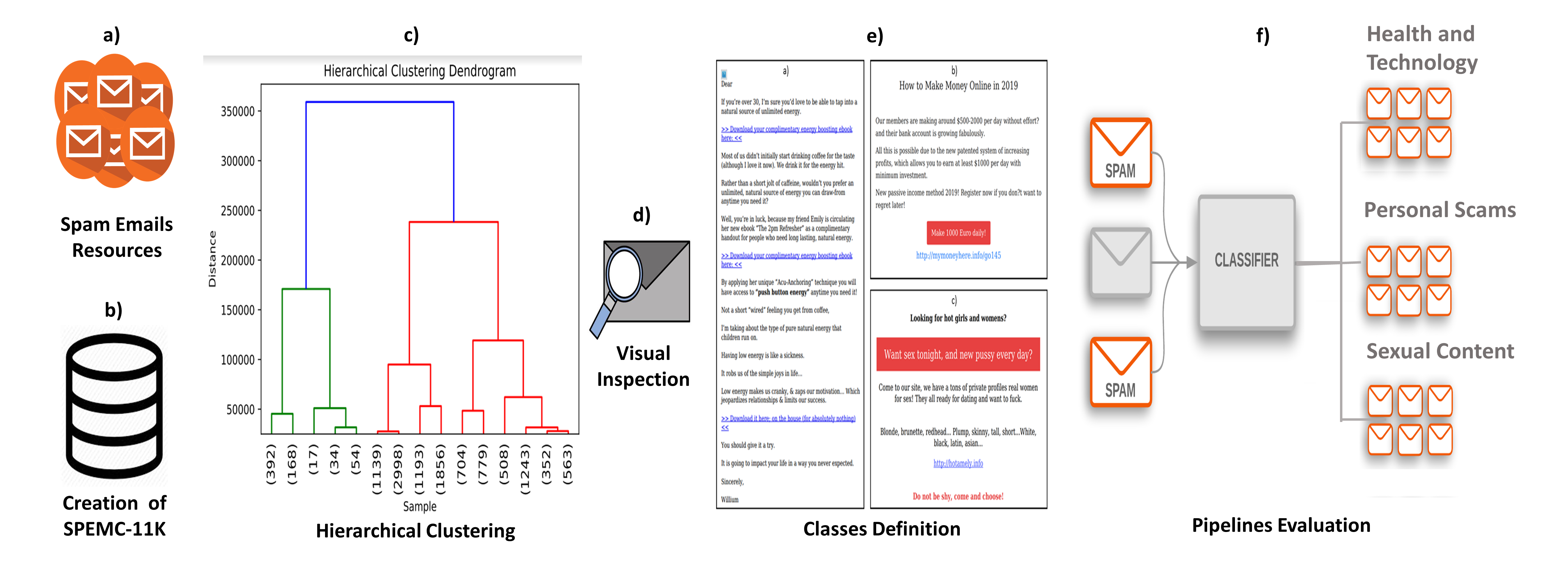}
    \caption{Spam email multi-classification process: a) 12.5K spam emails, b) pre-process and restrict to English language, c) hierarchical clustering, d) manual review of the clusters, e) category labelling, f) training and evaluation of six pipelines of text classification.}
    \label{fig:workflow}
\end{figure*}

To train ours multiclassification pipelines, we created a spam email multi-class dataset called Spam Email Classification $11K$ (SPEMC-$11$K). We evaluated six pipelines based on Text Classification techniques that combine two feature extractors, Term Frequency - Inverse Document Frequency (TF-IDF) and Bag of Words (BOW), and three Machine Learning algorithms, Support Vector Machine (SVM), Naive Bayes (NB), and Logistic Regression (LR). We followed the process shown in Figure \ref{fig:workflow}.

The rest of the paper is organised as follows: Section \ref{sec:state} presents the literature review about spam email challenges. Section \ref{sec:method} explains the methodology proposed to address the creation of SPEMC-$11$K and the set of the designed classification pipelines. After that, in Section \ref{sec:exp}, we discuss the experiments and results. Finally, Section \ref{sec:conclusions} presents our conclusion and our future work.  

\section{State of the Art}
\label{sec:state}

Spam email has been a problem for more than two decades since spammers compete against filter developers by creating and refining new dynamic and adaptive spammers techniques. In the recent years, many researchers investigated and presented new approaches to overcome this situation (\cite{Barushka_2018_SpamFilteringDL}, \cite{Bahgat_2018_EmailClassificationSematicMethods},  \cite{Faris_2019_RWNWithGASpamFilter}). Barushka and Hajek \cite{Barushka_2018_SpamFilteringDL} proposed a regularised deep multi-layer perceptron neural network as binary classifier and evaluated their model on the SpamAssassin dataset\footnote{\protect\url{https://spamassassin.apache.org/old/publiccorpus/} Retrieved March 2020} with $99.89\%$ of accuracy and the Enron-Spam dataset\footnote{\protect\url{http://nlp.cs.aueb.gr/software_and_datasets/Enron-Spam/index.html} Retrieved March 2020} with $98.76\%$ of accuracy. Bahgat et al. \cite{Bahgat_2018_EmailClassificationSematicMethods} evaluated their model based on a semantic feature selection with an SVM classifier on Enron-Spam Dataset reaching $94\%$ of accuracy. Faris et al. \cite{Faris_2019_RWNWithGASpamFilter} presented a binary model based on a Genetic Algorithm as a feature selector and the Random Weight Network as classifier obtaining $96.70\%$ of accuracy on the SpamAssassin dataset. 

Although these models achieved high performances on the most popular spam email datasets, these contain emails from the early $2000s$ and do not take into account the current spammer tricks. The variation of the not spam and spam emails over the time, namely \textit{concept drift}, was addressed by Ruano-Ordas et al. \cite{Ordas_2018_DriftEmailSpam} and discovered particular issues, such as the existence of topics associated to multiple forms of concept drift. Moreover, in another work, Ruano-Ordas et al. \cite{Ordas_2018_EvolutionaryComputationSpam} presented DiscoverRegex, a new regular expression finding tool that considers the drift concept, and Shujian et al. \cite{Shujian_2019_DriftDetectionHierchicalHypothesis} presented a framework to overcome the lack of drift detection on spam. 

The automatic classification of spam email into categories can help to handle the concept drift during a period by identifying spam classes patterns or detecting cybercrime campaigns, improving the traditional detection of spam emails. Besides, to the best of our knowledge, there are no works that tackle the spam email problem from a multi-classification perspective. We followed the intuition of works several works (\cite{Biswas_2020_PerceptualHashingTor, Alnabki_2019_ToRank, Biswas_2017_PerceptualHashingTOR, Fidalgo_2019_ClassifyingTorSemanticAttention, Alnabki_2017_ClassifyingIllegalActivities}), where authors tried to give a multi-classification overview of the content of Tor Darknet, instead of suggesting that Tor hosts legal or illegal content. 

\section{Methodology}
\label{sec:method}
\subsection{SPEMC-$11$K Dataset}

Based on our purpose to build a multi-classifier to categorise the spam emails, we randomly extracted $12,5K$ English spam emails collected by the Spanish National Cybersecurity Institute (INCIBE) between April and November 2019. Although the filters can classify spam based on processing the email's headers, body or the attachments, in our work we only used the email body, since we have focused on a text classification approach. 

First, to select only English emails, we used Langdetect \footnote{\protect\url{https://pypi.python.org/pypi/langdetect}}. Second, we preprocessed them by removing special characters, single letters, numbers, and stop-words. Then, following the methodology used by Biswas et al. \cite{Biswas_2020_PerceptualHashingTor}, we applied an unsupervised hierarchical clustering based on Ward's minimum variance. Next, after applying the hierarchical clustering, we discarded emails with less than five words, obtaining a set of $11462$ emails. Finally, we labelled three well-differential classes by merging clusters and identifying the topics of each one. The final multi-class dataset is named as SPEMC-$11K$ (Spam Email Classification Dataset).

\begin{figure}[ht]
    \centering
    \includegraphics[scale=0.28]{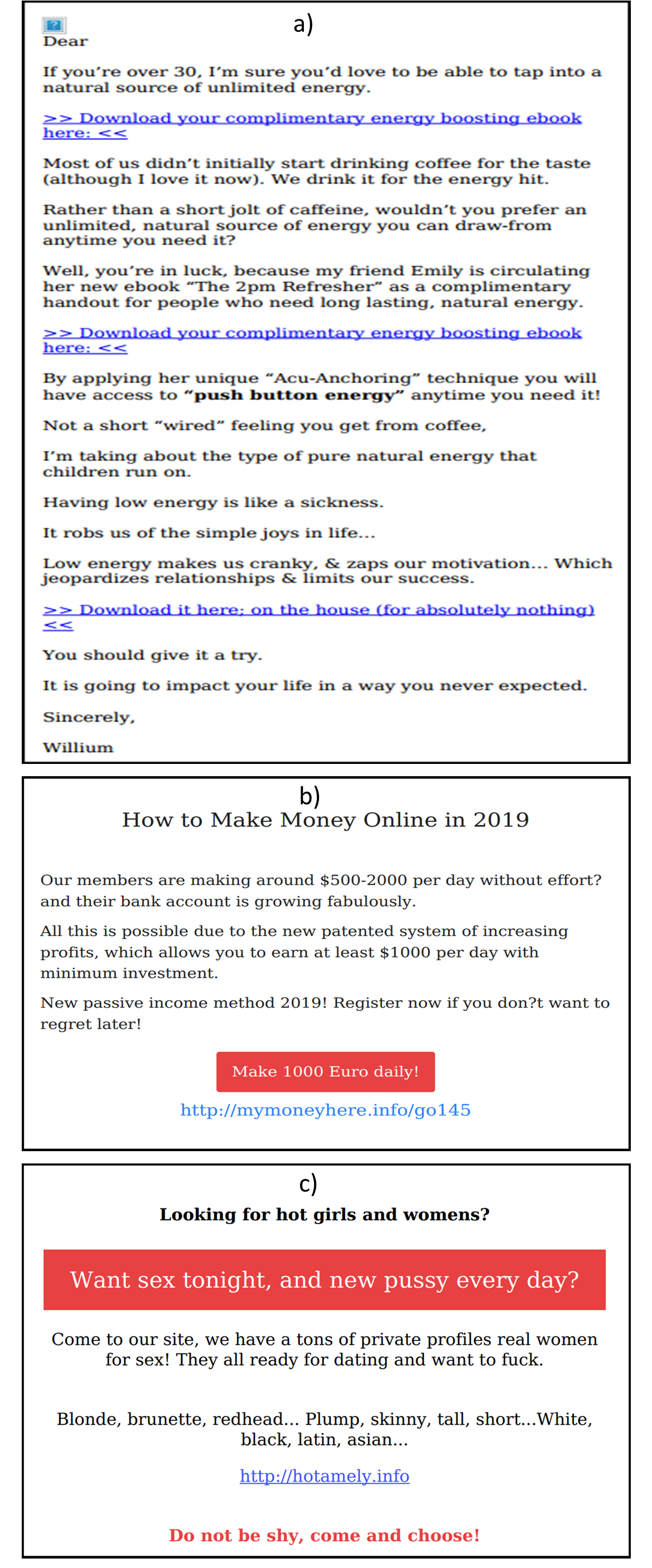}
    \caption{Spam email examples of the different classes: a) Health and Technology, b) Personal Scams, and c) Sexual Content.}
    \label{fig:categories}
\end{figure}

The Table \ref{tab:categories} shows the three spam classes, the number of examples per category, and their percentage in SPEMC-$11K$, and the Fig. \ref{fig:categories} shows spam emails examples of each category. \textbf{Health and Technology} category contains spam emails related to illegal sale of pills, miracle products, invitation to false conferences, and shocking news. \textbf{Personal Scams} class groups spam emails whose aim is to obtain illegally money or personal information from the users. \textbf{Sexual Content} category has spam emails with web sites of dates, pornography videos and photos, or text with sexual content. 
\begin{table}[ht]
\centering
\caption{Categories, number of examples per category, and percentage found in SPEMC-$11$K.}
\label{tab:categories}
\resizebox{\linewidth}{!}{
\begin{tabular}{c c c }
\hline
\textbf{Category}            & \textbf{Number of examples} & \textbf{Percentage (\%)} \\\hline\hline
Health and Technology & 583                         & 5.08                  \\
Personal Scams      & 3703                        & 32.31                 \\
Sexual Content     & 7176                        & 62.61                 \\
\textbf{Total}               & 11462                       &         \\\hline\hline
\end{tabular}
}
\end{table}

\subsection{Classification Pipelines}

We evaluated six classification pipelines, which are the combination of two encoding techniques with three supervised classifiers. As for the encoders, we considered two algorithms: (1) \textit{Bag of Words (BOW)} \cite{Harris_1954_bow} which extracts the features from the text corpus by counting the number of occurrences a word, and thus, creating a sparse feature vector where each element represents the occurrences of a single word; and (2) \textit{Term Frequency Inverse Document Frequency (TD-IDF)} \cite{Aizawa_2003_TFIDF} that builds a sparse vector based on a statistical model by assigning a numerical value for each word of a text corpus. TF-IDF emphasizes a word that frequently appears in a given text, whereas words that frequently occur in many texts of the corpus are penalized.
TF-IDF and BOW are straightforward and computationally efficient techniques to encode the text, although they do not consider the order or meaning of the word. 

To classify the vectorized text, we considered three classifiers: (1) Logistic Regression (LR) \cite{Cox_1958_LR}, (2) Support Vector Machines (SVM)  \cite{Vapnik_1995_SVM}, and (3) Naive Bayes (NB), which have been assessed as binary spam classifiers (\cite{Bhowmick_2018_SpamReviewLiterature}) due to their high performance. 
By combining each of the two vectorizers and the three classifiers, we got six combinations to compare over the SPEMC-$11$K dataset.

\section{Experimentation}
\label{sec:exp}

\subsection{Experimentation settings}

We used for this experimentation a laptop Dell Inspiron $5584$ with Intel(R) Core(TM) $i7-7500U$ CPU @ $2.70GHz$, $2901$ Mhz, $2$ main processors, $4$ logical processors with $16$G of RAM on Ubuntu $18.04$ OS and Python $3$ libraries. We applied Scipy\footnote{\protect\url{https://www.scipy.org/}} to made the hierarchical clustering. We generated SPEMC-$11$K, an unbalanced dataset where the Sexual Content category represents a $62.61\%$ of the total, followed by Personal Scams and Health and Technology, with $32.31\%$ and $5.08\%$ respectively. We implemented the six pipelines with Scikit-Learn\footnote{\protect\url{https://scikit-learn.org/stable/}}, and we managed the unbalanced data through a class-weight approach, assigning a weight related to the proportion of each class and its number of samples. 

We modified the models parameters in order to attempt to achieve the highest performance. For both the BOW and TF-IDF dictionary, we select a maximum of $9000$ words, and we considered only $1$-gram and a minimum number of word appearances of $3$ words in both text representations techniques. Regarding classifier parameters, we activated the class-weight and set a C value of $1000$ for LR. For SVM, We selected ``linear" as a kernel, C parameter to $1000$, and activated the class-weight flag. C parameter is an optimiser for both classifiers, where a high value looks for a lower margin of hyperplane separation. Finally, we chose a multinomial NB and kept their default parameter configuration. 

\subsection{Results and Discussion}

We calculated the macro, micro and weighted averages of the precision, recall, and F1 scores, the average accuracy on 5-fold cross-validation (CV), and the speed of execution per email as shown in Table \ref{tab:results}. 

The results showed a high overall performance in every pipeline, where the best accuracy value was achieved by the combination of TF-IDF and SVM with $95.39\%$ and F1-score value in a macro average of $95.39\%$. Also, we observed that in general, the TF-IDF models overcome the BOW combinations, being NB the classifier that offers the most notable difference with $87.68\%$ of accuracy. These lower results can be due that NB does not consider the relationship between features and classes. In addiction, both NB pipelines offers the fastest execution times per email with $2.13$ms along with TF-IDF and $5.52$ ms with BOW, outperforming the TF-IDF-SVM, $306.56$ms, BOW-SVM, $177.86$, pipelines. Hence, although NB pipelines obtained the lowest performances of accuracy, they are more recommended for a real-time application due to their rapidity. However, if quickness is not a vital characteristic, TF-IDF-SVM is the most adequate choice. 

\begin{table}[tb]
\centering
\caption{Results of the comparison between the six multi-classification pipelines evaluating on SPEMC-$11$K with respect to 5 folds cross-validation accuracy (CV), precision (P), recall (R) and F1 score (F1) metrics for micro, macro and weighted averaging. The last column shows the speed of execution per email.}
\label{tab:results}
\resizebox{\linewidth}{!}{
\begin{tabular}{c c c c c c c c}
\hline
\textbf{Metrics}   & \textbf{} & \textbf{Average} & \textbf{Average} & \textbf{Average}    & \textbf{CV}       & \textbf{Time (ms)} \\
\textbf{Methods}   &           & \textbf{(micro)} & \textbf{(macro)} & \textbf{(weighted)} & \textbf{Accuracy} & \textbf{per email} \\\hline\hline
\textbf{TFIDF-LR}  & P         & 0.9482           & 0.9099           & 0.9489              & 0.9482            & 48.64              \\
\textbf{}          & R         & 0.9482           & 0.9429           & 0.9482              &                   &                    \\
\textbf{}          & F1        & 0.9482           & 0.9251           & 0.9483              &                   &                    \\
\textbf{TFIDF-SVM} & P         & 0.9539           & 0.9094           & 0.9539              & \textbf{0.9539}   & 306.56             \\
\textbf{}          & R         & 0.9539           & 0.9538           & 0.9299              &                   &                    \\
\textbf{}          & F1        & \textbf{0.9539}  & \textbf{0.9539}  & \textbf{0.9545}     &                   &                    \\
\textbf{TFIDF-NB}  & P         & 0.9294           & 0.9064           & 0.9294              & 0.9294            & \textbf{2.13}      \\
\textbf{}          & R         & 0.9294           & 0.8534           & 0.8768              &                   &                    \\
\textbf{}          & F1        & 0.9294           & 0.9294           & 0.9275              &                   &                    \\
\textbf{BOW-LR}    & P         & 0.9472           & 0.9069           & 0.9531              & 0.9472            & 33.59              \\
\textbf{}          & R         & 0.9472           & 0.9649           & 0.9472              &                   &                    \\
\textbf{}          & F1        & 0.9472           & 0.9330           & 0.9483              &                   &                    \\
\textbf{BOW-SVM}   & P         & 0.9527           & 0.9401           & 0.9585              & 0.9527            & 177.86             \\
\textbf{}          & R         & 0.9527           & 0.9696           & 0.9529              &                   &                    \\
\textbf{}          & F1        & 0.9527           & 0.9529           & 0.9536              &                   &                    \\
\textbf{BOW-NB}    & P         & 0.8769           & 0.8779           & 0.8807              & 0.8769            & 5.52               \\
\textbf{}          & R         & 0.8769           & 0.7523           & 0.8769              &                   &                    \\
\textbf{}          & F1        & 0.8769           & 0.7801           & 0.8640              &                   &                   
\\\hline\hline\end{tabular}
}
\end{table}

\section{Conclusions}
\label{sec:conclusions}

To the best of our knowledge, in this work, for the first time in the literature, we dealt with the problem of spam detection through the use of multi-classification, instead of a binary classifier. 
We applied a hierarchical clustering on $12,5K$ English spam emails and discarded emails with less than five words, resulting in an unbalanced dataset of $11462$ emails divided into three categories: Health and Technology, Personal Scams, and Sexual Content, with $62.61\%$, $32.31\%$, and $5.08\%$ respectively. We used this data for evaluating the combination of two feature extractors, TF-IDF and BOW, along with three Machine Learning classifiers, LR, SVM, and NB. The combination of TF-IDF with SVM achieved the best accuracy, $95.39\%$. Instead, TF-IDF-NB achieved the fastest execution time per email, $2.13$ms being the most suitable pipeline to use in real-time applications. 

On the one hand, we have explored the content of spam emails as a multi-classification task for the first time. The results encourage us to divide the three spam classes into more classes and investigate the content of each one deeply to understand the spammers' behaviours. The research on the class Health and Technology, i.e., spam that contains products, could help to detect emerging products as Al Nabki et al. \cite{Alnabki_2017_DetectingEmergingProducts} did on the Darknet Tor. These can also help to anticipate or detect the concept drift in spam emails and define the use of spammers tricks over time. Moreover, considering another work of Al Nabki et al. \cite{Alnabki_2019_ToRank} could lead to developing a model to rank and detect the most influential hidden services inside spam emails.

On the other hand, the high performance obtained by the pipelines suggested that the problems related to the concept drift in binary classification can occur with the multi-classification. Hence, we plan to assess the six pipelines on public datasets and other spam email time sets provided by INCIBE.
 
\section*{Acknowledgement}

This work was supported by the framework agreement between the Universidad de Le\'{o}n and INCIBE (Spanish National Cybersecurity Institute) under Addendum 01.

\bibliographystyle{IEEEtran}
\bibliography{bibliography}

\end{document}